\documentclass[10pt,twocolumn,letterpaper]{article}

\usepackage{iccv}              

\newcommand{\hidecontent}[1]{}

\definecolor{iccvblue}{rgb}{0.21,0.49,0.74}
\definecolor{baselinecolor}{gray}{.9}
\definecolor{mygray}{rgb}{0.9, 0.9, 0.9}

\newcommand{\ours}{USP\xspace}
\usepackage{graphicx}
\usepackage{subcaption}

\definecolor{iccvblue}{rgb}{0.21,0.49,0.74}

\usepackage[pagebackref,breaklinks,colorlinks,allcolors=iccvblue]{hyperref}
\def\paperID{7499} 
\def\confName{ICCV}
\def\confYear{2025}

\usepackage{multirow}
\usepackage{xcolor,colortbl}
\usepackage{diagbox}
\usepackage{amsfonts}
\usepackage{collcell}

\title{USP: Unified Self-Supervised Pretraining for Image Generation and Understanding}

\author{Xiangxiang Chu ~~ Renda Li ~~ Yong Wang\\
AMAP, Alibaba Group\\
{\tt\small \{chuxiangxiang.cxx, lirenda.lrd, wangyong.lz\}@alibaba-inc.com
}}

\begin{document}
\maketitle

\begin{abstract}
Recent studies have highlighted the interplay between diffusion models and representation learning. 
Intermediate representations from diffusion models can be leveraged for downstream visual tasks, while self-supervised vision models can enhance the convergence and generation quality of diffusion models. 
However, transferring pretrained weights from vision models to diffusion models is challenging due to input mismatches and the use of latent spaces. To address these challenges, we propose Unified Self-supervised Pretraining (USP), a framework that initializes diffusion models via masked latent modeling in a Variational Autoencoder (VAE) latent space. USP achieves comparable performance in understanding tasks while significantly accelerating convergence and enhancing generation quality in diffusion models. Our code is available on \url{https://github.com/AMAP-ML/USP}.
\end{abstract}

\section{Introduction}
\label{sec:intro}
Over the past decade, the pretraining-finetuning paradigm has achieved remarkable success in image recognition. Strong representations are first obtained by pretraining neural networks on large-scale datasets \cite{he2016deep,szegedy2015going,huang2016deep,dosovitskiy2020image,chu2021twins,liu2021swin,he2020momentum,he2022masked,caron2021emerging, chu2024visionllama}. These pretrained models are then fine-tuned for downstream tasks such as object detection \cite{ren2015faster,redmon2016yolo,he2017mask, tian2019fcos} and segmentation \cite{long2015fully,xiao2018unified,xie2021segformer,zhang2022segvit}, achieving state-of-the-art performance. Meanwhile, self-supervised learning has also seen rapid development, with methods such as \cite{he2020momentum,chen2020simple,he2022masked,bao2021beit} catching up to and even surpassing their supervised counterparts \cite{he2016deep, dosovitskiy2020image, liu2021swin, chu2021twins,chu2023conditional,touvron2022deit}.

In parallel, image generation has also experienced significant progress, though following a different trajectory.
Diffusion models \cite{dhariwal2021diffusion,Song2020,Song2020b} have emerged as a dominant force, outperforming Generative Adversarial Networks (GANs) \cite{goodfellow2014generative} and opening a new era in generative modeling. Previous studies \cite{donahue2019large,amit2021segdiff,wolleb2022diffusion,wei2023diffusion,brempong2022denoising,baranchuk2022labelefficient, xiang2023denoising} have shown that generative models can learn competitive recognition representations, indicating a strong relationship between image understanding and generation.

Pioneer works, such as iGPT \cite{chen2020generative}, have explored autoregressive pretraining in the pixel space. 
However, this approach faces substantial challenges when scaling to large datasets and models. Furthermore, it is inherently incompatible with diffusion models.
To address these limitations, a recent study, REPA \cite{yu2025repa}, proposes aligning the representations of diffusion models with those from off-the-shelf pretrained recognition models. By aligning intermediate-layer representations, diffusion models can more efficiently acquire discriminative features, leading to faster convergence and improved performance within the same computational budget.
However, this method relies on powerful pretrained backbones, such as DINOv2 \cite{oquab2024dinov}, which require substantial data and computational resources (over 22,000 A100 GPU hours). Additionally, the extra teacher backbone not only increases GPU memory consumption but also reduces the training speed of diffusion models.

Several fundamental questions remain unresolved: Is pretraining feasible and necessary for diffusion-based image generation? Can we establish a unified pretraining framework that simultaneously improves image generation and state-of-the-art perception tasks—in other words, a single pretraining paradigm beneficial to both image generation and recognition? Can the widely-adopted pretraining-finetuning paradigm also succeed in generative models? This paper aims to address these questions by proposing a simple yet effective approach.

A major challenge in this endeavor is the input mismatch between perception and generation models, as recently noted in literature \cite{yu2025repa}. Specifically, perception backbones typically process clean images, whereas diffusion models inherently handle images perturbed by scheduled noise. This discrepancy is further complicated by modern diffusion models \cite{peebles2023scalable,rombach2022high,esser2024scaling} operating iteratively in a latent space mediated by VAEs.

To bridge this gap, we propose a two-step strategy. Our approach is grounded in a simple observation and a straightforward intuition: representation learning is central to various vision tasks, including image generation, so why not focus on enhancing it rather than being distracted by the nuances of downstream tasks?

Initially, we encode images into latent space using an off-the-shelf VAE. Subsequently, we perform masked latent modeling during pretraining, enhancing representation quality without labels or task-specific losses. This allows training a single model applicable across multiple downstream tasks, such as classification, segmentation, object detection, and image generation. With careful weight initialization, our pretrained model seamlessly integrates into downstream tasks. This approach not only promotes robust representation learning for both image recognition and generation but also eliminates the additional computational overhead associated with fine-tuning downstream tasks.

Our contributions are summarized as follows: 
\begin{itemize}
    \item  Inspired by the success of the pretraining-fine-tuning paradigm in image understanding, we propose USP, a unified pretraining framework that synergizes image comprehension and diffusion-based image generation. 
    
    \item  To decouple pretraining from the heterogeneous optimization objectives of downstream tasks, we propose masked feature modeling within the latent space of VAEs. This approach enables robust representation learning in an unsupervised manner, eliminating the need for labeled data and facilitating fast training.
    
    \item Our approach, through meticulous weight initialization, significantly enhances the performance of two transformer-based diffusion models: DiT and SiT. It achieves superior results compared to their baselines—trained for 7M steps—in just 600K and 150K steps, respectively, yielding speedups of 11.7$\times$ and 46.6$\times$. Importantly, it maintains strong representation capabilities for image recognition, underscoring its versatility. Additionally, it incurs no extra training cost or memory overhead, ensuring high efficiency and scalability. Moreover, our method is somewhat orthogonal to other acceleration methods such as REPA \cite{yu2025repa} and VAVAE \cite{vavae}.
\end{itemize}
\section{Method}
\label{method}
\subsection{Motivations}
At first glance, several challenges appear to hinder the successful implementation of a unified pretraining-fine-tuning paradigm for both image recognition and generation.
\begin{itemize}
    \item C1: \textbf{Input Mismatch}. Image recognition model accepts clean images while diffusion model takes noisy images as input. 
    \item C2: \textbf{Architecture Mismatch}. Modern generative models are latent diffusion models using VAE while VAE is rarely used in image understanding tasks. 
    Moreover, the common ViT architecture in image comprehension \cite{dosovitskiy2020image} has been modified in image generation \cite{peebles2023scalable}.
    \item C3: \textbf{Divergent Loss Functions and Labels.} Image recognition and generation tasks often employ different loss functions and label formats, complicating the integration of these tasks within a unified framework.
\end{itemize}
However, several promising observations provide a foundation for overcoming these challenges.

\begin{itemize}
    \item P1: \textbf{Robustness of Neural Networks}. 
    Neural networks demonstrate robustness to data noise, with pretrained backbones \cite{hendrycks2018benchmarking,oquab2024dinov,he2022masked,liu2022convnet,chen2024revealing}  maintaining significant accuracy on noisy or augmented datasets like ImageNet-C \cite{hendrycks2018benchmarking}. This benchmark \cite{hendrycks2018benchmarking} closely resembles the noise injection process in diffusion models, making it a relevant test for model robustness.
    
    \item P2: \textbf{Discriminative Representations in Diffusion Models.} Diffusion models have been shown to learn discriminative representations \cite{amit2021segdiff, baranchuk2021labelefficient,mukhopadhyay2024text,xiang2023denoising}. Aligning diffusion models with pretrained visual encoders can significantly enhance convergence and final performance \cite{yu2025repa}. Notably, the enhanced linear probe classification of intermediate layers plays a crucial role in boosting system-level performance. 

    \item P3: \textbf{Adaptability of Modified ViT Architectures}. As discussed in the latter part of C2, adapting the original ViT architecture to diffusion models requires modifications, including AdaLN-Zero and conditional inputs (e.g., class labels and timesteps). Nevertheless, through carefully designed strategies (detailed in Sec~\ref{sec:init adaption}), the modified ViT remains compatible with pretrained ViT weights.
    
    \item P4: \textbf{Strong Compression and Reconstruction Capabilities of VAEs}. Off-the-shelf VAEs used in diffusion models exhibit robust compression and reconstruction abilities, effectively preserving the essential information of the original images with minimal loss \cite{esser2024scaling}.
\end{itemize}

\begin{figure*}[h]
  \centering
  \includegraphics[width=\linewidth]{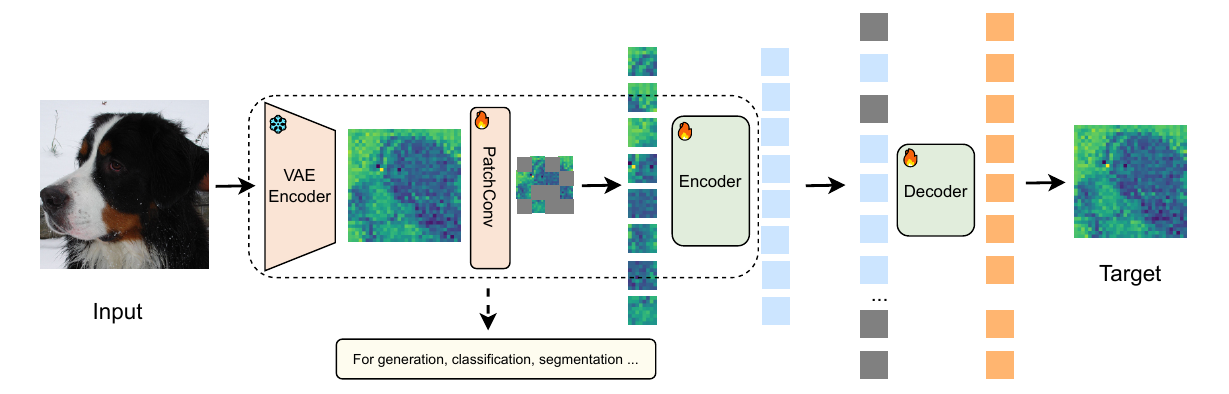}
  \caption{\textbf{Overall architecture.} The input image is downsampled by SD-VAE and PatchConv to generate image patches. A subset of these patches is randomly masked according to a predefined ratio. The encoder processes the unmasked patches, while the decoder reconstructs the masked ones. During pretraining, VAE parameters are frozen. In the post-pretraining, PatchConv and encoder weights \textbf{initialize} downstream tasks for understanding and generation, and the decoder is \textbf{discarded}.
}
  \label{fig:method}
\end{figure*}

Based on the above discussions, we propose a multi-faceted strategy to address the identified challenges. We project the image into its latent space using the frozen off-the-shelf VAE used in generative models to address the first half of C2, which can benefit from the advantage of P4. Moreover, we can address the remaining issue of C2 by using curated adaptations in P3. Instead of solving C3 directly, we take an immediate alternative. Drawing inspiration from P2, we design our training objective to foster robust discriminative representations, essential for both image understanding and generation tasks. This approach effectively circumvents the challenges posed by divergent loss functions and labels, thereby harmonizing the training process across both domains.
Combining P1, strong representations can greatly alleviate the influence of C1. Furthermore, we carefully calibrate the normalization to increase the matching degree of input data distribution. 

\subsection{Model Architecture}
The overall pipeline of \ours is illustrated in Figure~\ref{fig:method}. \ours is based on a simple autoencoder framework, similar to that of \cite{he2022masked}, where we reconstruct masked feature patches from the unmasked ones in the latent space encoded by a VAE. We opted not to use contrastive learning methods  \cite{he2020momentum,chen2020simple,grill2020bootstrap,caron2020unsupervised} due to their lower training efficiency, although our approach could also be adapted to these frameworks.

Given an input image $I \in \mathbb{R}^{H \times W \times 3}$, its latent feature $I_{vae}$ is obtained through a frozen VAE. For example, using the SD-VAE \cite{rombach2022high} with $H=W=224$, we obtain a compressed feature $I_{vae} \in \mathbb{R}^{\frac{H}{8} \times \frac{W}{8}\times 4}$. This latent feature is then divided into non-overlapping patches using 2$\times$2 convolutions (referred to as PatchConv), resulting in patches of size  $\frac{H}{16}\times \frac{W}{16} \times h$, comparable to those used in \cite{dosovitskiy2020image}. A class token is added to the image tokens and enhanced with their sine-cosine position encodings. These patches are randomly masked with a specified mask ratio $m$. The unmasked patches are first processed by a standard ViT encoder \cite{dosovitskiy2020image} with $h$ hidden dimensions, and then concatenated with the masked patches as input to an asymmetric vision decoder for reconstruction. After pretraining, the decoder is discarded, and only the ViT encoder and PatchConv are retained. Finally, the weights of the pretrained model are used to initialize downstream tasks such as classification, semantic segmentation, and generation. Therefore, this approach doesn't introduce either extra training cost or memory overhead for downstream tasks.

\textbf{Data Augmentation and Normalization.} 
We adopt a weak data augmentation strategy, specifically RandomHorizontalFlip. This approach offers two key advantages: 1) it reduces the data augmentation gap between pretraining and generation; 2) it allows for the use of caches to accelerate training. 
We utilize the normalization scheme of the VAE, instead of the widely used ImageNet normalization. Although this modification deviates from zero mean and unit variance, it ensures consistency in the data distribution between pretraining and fine-tuning.

\textbf{Loss Function.} 
We employ the mean squared error (MSE) loss between network outputs and masked patches in latent space, with each patch individually normalized.

\subsection{Initialization Adaption}
\label{sec:init adaption}
For image recognition tasks, the pretrained weights can be seamlessly inherited without additional effort. The class token is reused for classification. 

For generative tasks, we carefully adjust the network architecture to ensure consistency between the pretrained ViT model and the initialized DiT/SiT models. Specifically, we introduce slight modifications to the AdaLN-Zero layer normalization scheme. We reintroduce trainable bias $\beta$ and scale $\gamma$  parameters to inherit the weights of the pretrained backbone fully. This adjustment represents a negligible increase in the total number of trainable parameters of DiT/SiT. Given that our model is pretrained at a resolution of 224$\times$224, we employ bicubic interpolation to upscale the positional embeddings from 224$\times$224 to 256$\times$256 for ImageNet generation tasks at the 256$\times$256 resolution. The class token is directly abandoned. 
\section{Experiments}
\label{sec:experiments}
\subsection{Pretraining}
Our paper covers three models, with detailed settings presented in Table~\ref{tab:model_setting}. 
All models share the same decoder—consisting of 8 Transformer blocks with a hidden size of 512—and are trained for 800 epochs with a default global batch size of 4096 to enable fair comparisons and ablation studies.

\begin{table}[h]
  \centering
  \begin{tabular}{@{}lccc@{}}
    \toprule
    Model & Layers & Heads & Dims  \\
    \midrule
    Base & 12 & 12 & 768 \\
    Large & 24 & 16 & 1024\\
    XL & 28 & 16 & 1152\\
    \bottomrule
  \end{tabular}
  \caption{Model settings.}
  \label{tab:model_setting}
\end{table}
We extend the training duration to 1600 epochs in an additional experiment to demonstrate the scalability of \ours with respect to pretraining time. We use AdamW \cite{loshchilov2017decoupled} optimizer with $\beta_1=0.95$ and $\beta_2=0.9$.  The learning rate is warmed up to $2.4\times$10$^{-3}$ in the first 40 epochs and subsequently decayed to zero following a cosine strategy. We apply a weight decay of 0.05 to regularize the model.

To ensure a fair comparison with MAE \cite{he2022masked}, we pretrain at a resolution of 224$\times$224, although experiments indicate that higher performance can be achieved with larger resolutions. We utilize a mask ratio of 0.75 and apply patch-wise normalization loss as MAE \cite{he2022masked}. The VAE is kept frozen throughout this paper. 
To maintain comparable training efficiency with MAE, we use the VAE to generate image latent features and construct caches in an off-the-shelf manner. 
Note that we adopt a different image normalization setting, with mean=[0.5, 0.5, 0.5] and std=[0.5, 0.5, 0.5], instead of the default ImageNet normalization.

\begin{figure*}[h]
  \centering
  \includegraphics[width=\linewidth]{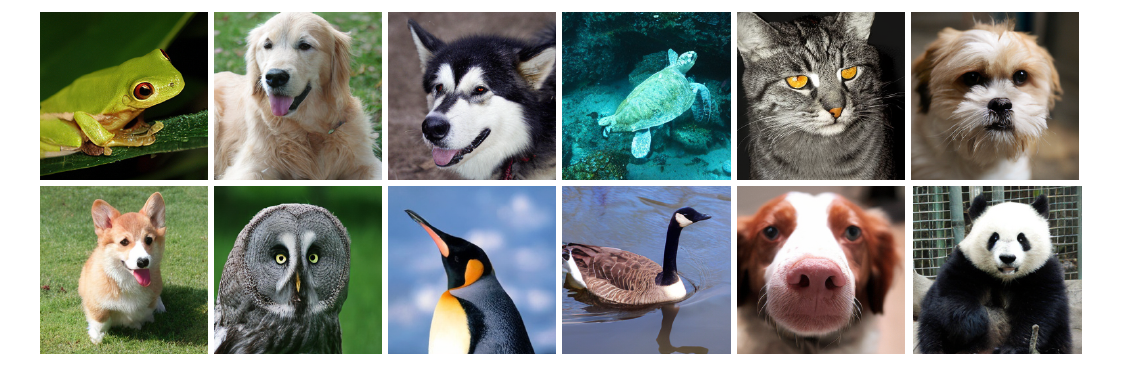}
  \vspace{-5mm}
  \caption{256$\times$256 generation samples: DiT-XL/2 (1.2M steps) with CFG=4.0.}
  \label{fig:image_gen_ditx2}
\vspace{-2mm}  
\end{figure*}

\subsection{Image Generation}
We validate our method on two transformer diffusion models: DiTs \cite{peebles2023scalable} (based on DDPM \cite{ho2020denoising}) and SiTs \cite{ma2024sit} (based on flow models \cite{lipman2022flow}. We adhere to the default training settings \cite{peebles2023scalable, ma2024sit}. For SiTs, we use linear interpolants and velocity predictions. Evaluation metrics include Fréchet Inception Distance (FID \cite{heusel2017gans}), Inception Score (IS \cite{salimans2016improved}), Precision (Pre.), and Recall (Rec.)\cite{kynkaanniemi2019improved}. Evaluations are conducted on ImageNet 256$\times$256 using 50K samples, without classifier-free guidance (CFG).

\subsubsection{Comparison Under the DiT Framework}
In general, we deliberately exclude certain enhancements to DiT, such as architecture modifications \cite{chu2024visionllama,gao2023masked}, advanced training frameworks \cite{lipman2022flow, liu2022flow,esser2024scaling}, and schedule refinements \cite{esser2024scaling}. These improvements are orthogonal to our approach and do not directly impact the core objectives of our study.

As shown in Table \ref{tab:comp_orig_dit}, \ours significantly improves performance across various DiT model sizes compared to corresponding baselines. Additionally, our method consistently improves generation quality as training progresses. For example, with the DiT-B/2 model, our approach achieves an FID of 24.18 after 1M training steps, outperforming the baseline by 31.26. Compared with a recent DiT variant \cite{chu2024visionllama}, which introduces a novel architecture requiring 2.5M steps, our method achieves better FID within only 400K steps.

\begin{figure}[h]
    \centering
    \includegraphics[width=\linewidth]{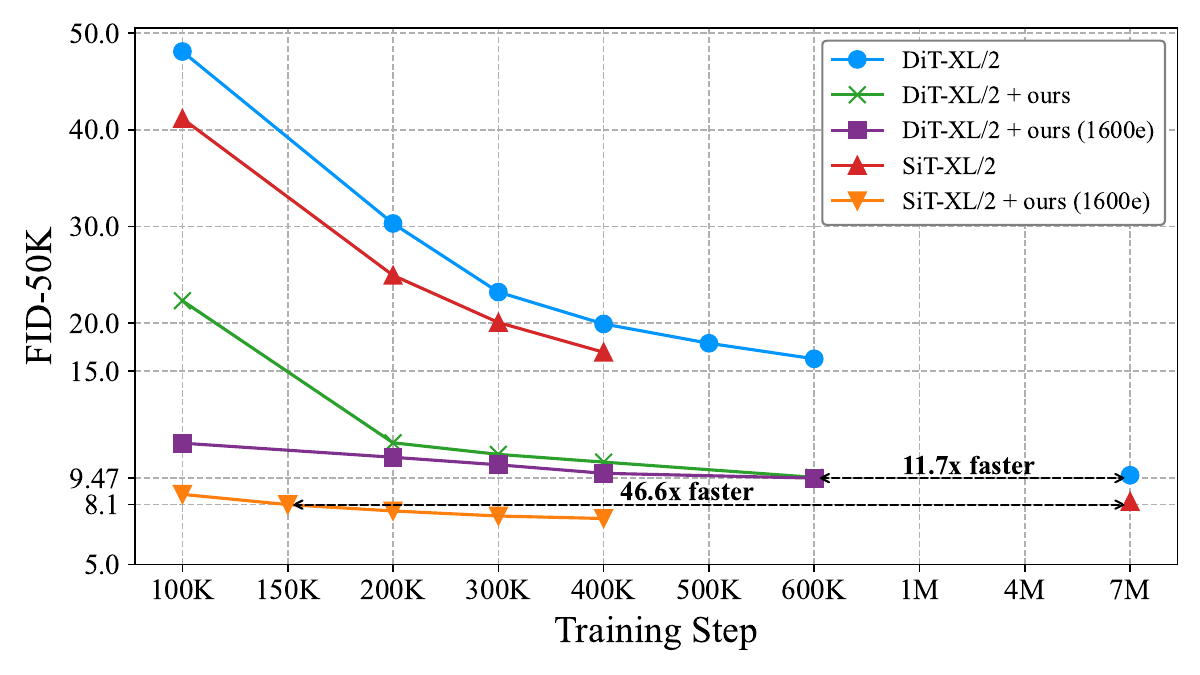}
    \caption{\textbf{Convergence acceleration.} 
    \ours accelerates the convergence of DiT and SiT by over \textbf{11.7×} and \textbf{46.6×}, respectively, and longer pre-training consistently leads to improved performance.}
    \label{fig:acc_dit}
\end{figure}

\begin{table}[h]
    \centering
    \resizebox{1.0\linewidth}{!}{
    \begin{tabular}{l c c c c }
        \toprule
        \textbf{Model} & \textbf{Params} & \textbf{Steps} & \textbf{FID} $(\downarrow)$ & \textbf{IS} $(\uparrow)$ \\
        \midrule
        DiT-B/2 & 130M & 400K & 42.62 & 33.67 \\
        \rowcolor{mygray}\ours 800/1600 & 130M & 400K & 28.26/27.1 & 48.92/50.46 \\
        DiT-B/2 & 130M & 1M & 31.26 & 47.35 \\
        \rowcolor{mygray}\ours & 130M & 1M & 24.18 & 58.05 \\
        
        \midrule
        DiT-L/2 & 458M & 400K & 23.03 & 60.13\\
        \rowcolor{mygray}\ours 800/1600 & 458M & 400K & 16.23/15.04 & 76.19/81.14 \\
        \midrule
        DiT-XL/2 & 675M & 400K & 19.94 & 67.16 \\
        \rowcolor{mygray}\ours 800/1600 & 675M & 400K & 10.28/9.73& 104.36/111.80 \\
        \midrule
        DiT-XL/2$^\dagger$ & 675M & 2.5M & 10.67  & - \\
        DiT-XL/2 & 675M & 7M & 9.62 & 121.50 \\
        DiT-LLaMA-XL/2  &675M & 2.5M & 9.84  &117.72 \\
        DiT-LLaMA-XL/2$^{\ddagger}$ &675M & 2.5M & 2.42 & 265.39\\
        \rowcolor{mygray} \ours & 675M & 1.2M & 8.93 & 123.10\\
        \rowcolor{mygray}\ours$^{\ddagger}$ & 675M & 1.2M & 2.33 & 267.07 \\

        \bottomrule 
    \end{tabular}
    }
        \caption{Comparison with  DiT and one of its leading variant. $\dagger$: result from \cite{chu2024visionllama}. $\ddagger$: with CFG. 800/1600: pretraining epochs.}
    \label{tab:comp_orig_dit}
\end{table}

We present the image generation samples in Figure~\ref{fig:image_gen_ditx2} and provide additional results in Section~\ref{sec:image_gen_256}.

\subsubsection{Comparison Under the SiT Framework} 
We further evaluate \ours on the SiT framework, using pretraining weights from 1600 epochs by default. As shown in Table \ref{tab:comp_repa_dit_sit}, \ours demonstrates consistent performance on the flow-based diffusion model. We present the image generation samples in Section~\ref{sec:image_gen_256}.

\textbf{Convergence Acceleration.}
We visualize the evolution of FID scores during training for vanilla DiT/SiT and our method, as shown in Figure~\ref{fig:acc_dit}. Our method achieves \textbf{11.7×} faster convergence on DiT-XL/2 and \textbf{46.6×} faster on SiT-XL/2. This advantage is more pronounced with extended pretraining, highlighting the effectiveness of our initialization strategy for generative tasks.

\subsubsection{Comparison with Acceleration Methods} 
We compare \ours with recent methods utilizing representation alignment to enhance DiT performance. As shown in Table \ref{tab:comp_repa_dit_sit}, \ours consistently achieves superior performance across various DiT model sizes. Notably, prior approaches such as \cite{yu2025repa} rely on powerful pretrained backbones (e.g., DINOv2), requiring extensive data and computational resources, whereas our method achieves superior performance more efficiently.

\subsubsection{Orthogonality to External-Model-Based Methods}
Although \ours achieves significantly faster convergence for diffusion models without relying on any external models, we further verify its orthogonality to recent DINO-based acceleration methods~\cite{yu2025repa, vavae}. As shown in Table~\ref{tab:orth}, combining \ours with either VAVAE~\cite{vavae} or REPA~\cite{yu2025repa} leads to even faster convergence than using any single method alone.

\begin{table}[h]
    \centering
    \resizebox{0.9\linewidth}{!}{
    \begin{tabular}{l c c c c }
        \toprule
        \textbf{Model} & \textbf{Params} & \textbf{Steps} & \textbf{FID} $(\downarrow)$ & \textbf{IS} $(\uparrow)$ \\
        \midrule
        SiT-XL/2 & 130M & 400K & 16.97 & 77.50 \\
        \ours & 130M & 400K & 7.38 & 127.96 \\
        REPA & 130M & 400K & 7.9 & 122.6 \\
        \ours + REPA & 130M & 400K & 6.26 & 139.84 \\
        \midrule
        VAVAE & 130M & 64 Epochs  & 5.18/2.15$^\dagger$ & 132.4/245.1$^\dagger$ \\
        \midrule
        \ours + VAVAE & 130M & 64 Epochs & 4.2/1.81$^\dagger$ & 144/261.0$^\dagger$ \\
        \bottomrule 
    \end{tabular}
    }
        \caption{Results Combined with External-Model-Based Methods. $^\dagger$: w CFG=10.0.}
        
    \label{tab:orth}
\end{table}

\begin{table}[h]
    \centering
    \resizebox{\linewidth}{!}{
    \begin{tabular}{l c c c c c}
        \toprule
        \textbf{Model} & \textbf{Method} & \textbf{Params} & \textbf{Steps} & \textbf{FID} $(\downarrow)$ & \textbf{IS} $(\uparrow)$ \\
        \cmidrule(lr){1-6}

        \multirow{3}{*}{DiT-B/2}
        & ReaLS (DINOv2-L) \cite{reals} & 130M & 400K & 35.27 & 37.80  \\
        & EQ-VAE \cite{kouzelis2025eq} &130M& 400K&34.1 &-\\
        \cmidrule(lr){2-6}
        \rowcolor{mygray}& \ours & 130M & 400K & 27.1 & 50.46  \\
        \cmidrule(lr){1-6}
        \multirow{2}{*}{DiT-L/2}
        & REPA \cite{yu2025repa} (DINOv2-B) & 458M & 400K & 15.6 & -  \\
        \cmidrule(lr){2-6}
        \rowcolor{mygray}& \ours  & 458M & 400K & 15.04 & 81.14 \\
        \cmidrule(lr){1-6}
        
        \multirow{3}{*}{DiT-XL/2}
        & REPA (DINOv2-B) & 675M & 400K & 12.3 & -  \\
        & EQ-VAE \cite{kouzelis2025eq} & 675M & 400K & 14.5 & 81.5\\
        \cmidrule(lr){2-6}
        \rowcolor{mygray}& \ours & 675M & 400K & \textbf{9.73} & 111.8 \\
        \midrule

        \multirow{3}{*}{SiT-B/2}
        &SiT-B/2 & 130M & 400K & 35.29 & 42.27 \\
        & ReaLS (DINOv2-L) & 130M & 400K & 27.53 & 49.70  \\
        & EQ-VAE &130M & 400K & 31.2 &- \\
        & REPA (DINOv2-B) & 130M & 400K & 24.4 & 43.7  \\
        & REPA (MAE-L)$^\dagger$ & 130M & 400K & 31.96  & 46.86 \\

        \rowcolor{mygray}& \ours & 130M & 400K & 22.10 & 61.20  \\
        \cmidrule(lr){1-6}
        \multirow{4}{*}{SiT-XL/2} 
        &         SiT-XL/2 & 675M & 400K & 16.97 & 77.50 \\
        & ReaLS  & 675M & 400K &  14.24 & 83.83  \\
        & EQ-VAE & 675M & 400k &16.1 &79.7  \\
        & REPA (DINOv2-B) & 675M & 400K & 7.9 & 122.6  \\
        \rowcolor{mygray} & \ours & 675M & 400K & \textbf{7.38} & 127.96 \\   
        \rowcolor{mygray} & \ours & 675M & 2M & 6.96 & 141.48 \\
        \bottomrule 
    \end{tabular}
    }
    \caption{Comparison with recent acceleration methods on DiTs and SiTs.  $\dagger$: reproduced using the code from \cite{yu2025repa}.}
    \vspace{-4mm}
    \label{tab:comp_repa_dit_sit}
\end{table}

\subsubsection{Training Cost}
We compare the training cost at each stage and present the results in Table~\ref{tab:training_cost}. Compared to the original DiT and REPA, our approach achieves a threshold FID of 9.6 using significantly fewer computational resources (approximately 15$\%$ and 5$\%$, respectively). Notably, our pretrained model can also be utilized for downstream tasks such as image classification, segmentation, and object detection. In this context, the pretraining cost can be further amortized across multiple tasks, enhancing the overall efficiency of our approach. 

With comparable computational costs, we retrained SiT-B/2 using MAE-ViT-Large for 400k steps. The results are summarized in Table~\ref{tab:comp_repa_dit_sit}. Our method achieved an FID of 22.1 and an IS of 61.2, significantly outperforming REPA, which reported an FID of 31.96 and an IS of 46.86.

\begin{table}[h]
  \centering
  \resizebox{0.9\linewidth}{!}{%
  \begin{tabular}{@{}lcccc@{}}
    \toprule
    Setting & FID &Total Cost & Pretrain & Fine-tuning   \\
    \midrule
    From scratch 7M & 9.6  & 622 & 0 & 622\\
    Pre 800 + FT 700K & 9.51& 98 & 36 & 62\\
    REPA FT 850K & 9.6& 1904 &1833$^{\dagger}$ & 71\\
    \bottomrule
  \end{tabular}
  }
  \caption{Training cost measured on DiT-XL/2 (in H20 GPU days). $^\dagger$: estimated from \cite{oquab2024dinov} based on our test.}
  \vspace{-4mm}
  \label{tab:training_cost}
\end{table}

\subsection{Image Understanding}
\subsubsection{Classification on ImageNet}
We conduct fine-tuning and linear probe classification experiments on the ImageNet-1k dataset. Specifically, we initialize the classifier with pretrained weights and freeze the VAE. We adhere strictly to the training settings described in \cite{he2022masked,chu2024visionllama}. All models are trained at an input resolution of 224$\times$224 and evaluated based on Top-1 accuracy on the validation set of ImageNet. The results are summarized in Table~\ref{tab:imagenet}. Despite the suboptimal hyperparameters for our method, it achieves comparable fine-tuning performance and significant improvements in linear classification.
Unlike linear probe, fully tuning transformer models demands robust data augmentations \cite{touvron2022deit,cubuk2020randaugment, yun2019cutmix,cubuk2019autoaugment,zhang2017mixup}. These augmentations, meticulously designed in the image space, are essential for enhancing model generalization and performance. However, the optimal tuning of SFT hyperparameters within the VAE-formed latent space, while significant, is not the primary focus of this paper.

\begin{table}[h]
  \centering
  \begin{tabular}{@{}lccc@{}}
    \toprule
    Method & Epochs &SFT Acc(\%) & LP Acc(\%) \\
    \midrule
    SemMAE \cite{li2022semmae} & 800 &83.4 & 65.0 \\
    SimMIM \cite{xie2022simmim} & 800 &83.8&56.7\\
    ViT-Base-MAE $^\ddagger$  & 800& 83.3 & 65.1\\
    \rowcolor{mygray} ViT-Base-\ours & 800 & 83.2 & 66.9 \\
    \midrule
    ViT-Large-MAE $^\ddagger$ & 800 & 85.4 & 73.7\\
    \rowcolor{mygray} ViT-Large-\ours & 800 & 84.8  & 74.5 \\
    \bottomrule
  \end{tabular}
  \caption{Classification result on ImageNet validation dataset. The VAE is the same as \cite{li2024autoregressive}. $\ddagger$: result from \cite{2023mmpretrain}.}
  \label{tab:imagenet}
\end{table}

We also conducted fine-tuning experiments using ViT-B with a trainable VAE, maintaining the same training settings. This model achieved a Top-1 accuracy of 81.3$\%$, indicating that our superior performance is not attributable to the increased computational cost (FLOPs) and parameter count of the frozen VAE.

\subsubsection{Semantic Segmentation on ADE20K}
We further evaluate our method on the semantic segmentation task using the MMSegmentation toolbox \cite{contributors2020mmsegmentation}. To make fair comparisons, we utilize the UpperNet framework \cite{xiao2018unified} and only replace our pre-trained backbone (and the frozen VAE) and strictly follow the training setting as \cite{chu2024visionllama,chu2021twins}. All models are trained for 160k steps with a global batch size of 16. We utilize the AdamW \cite{loshchilov2017decoupled} optimizer with $\beta_1=0.9$ and $\beta_2=0.999$. To avoid overfitting, we make use of weight decay of 0.05 and drop path rate 0.1. The initial learning rate is 0.0001 and decays to zero by a polynomial schedule. We report the single scale mIoU in Table~\ref{tab:ade20k}. Our method outperforms MAE by about 0.5$\%$ mIoU. 
\begin{table}[h]
  \centering
  \begin{tabular}{@{}lcc@{}}
    \toprule
    Method & Epochs &mIoU(\%) \\
    \midrule
    MAE $\dagger$ & 800& 46.2\\
    \ours & 800 & 46.7 \\
    MAE \cite{he2022masked} & 1600& 48.1 \\
    MaskFeat \cite{wei2022masked} & 1600 & 48.3\\
    \rowcolor{mygray}\ours & 1600 & 48.6\\
    \bottomrule
  \end{tabular}
  \caption{Segmentation result on ADE20k using ViT-Base. $\dagger$: results from \cite{chu2024visionllama}.}
  \vspace{-4mm}
  \label{tab:ade20k}
\end{table}
\subsection{Ablation Study}
In this section, we conduct an ablation study, with further details provided in Section~\ref{sec:ablation_more} (appendix).

\textbf{Role of VAE.}
As shown in Table \ref{tab: ablation_vae}, directly loading MAE’s pretrained weights—regardless of whether the Layer Norm (LN) parameters are used—yields inferior performance compared to training DiT from scratch. We hypothesize that this is due to these weight-loading strategies not preventing the downsampling convolution weights in DiT from being randomly initialized. This mismatch between the input distribution at the start of training and that during pretraining diminishes the effectiveness of the pretrained weights, leading to worse performance than training from scratch.

\begin{table*}[ht]
\scriptsize
\centering
\begin{minipage}[t]{0.3\textwidth}
    \begin{subtable}[t]{\linewidth}
\centering
\caption{\textbf{Role of VAE}}
\label{tab: ablation_vae}
\setlength\tabcolsep{1pt}
  \begin{tabular}{@{}lcccc@{}}
    \toprule
    Model & Pretrain & \textbf{FID} $(\downarrow)$  & \textbf{IS} $(\uparrow)$ \\
    \midrule
    DiT-B/2 orig. & from scratch & 42.62 &  33.67 \\
    \midrule
    DiT-B/2 orig. & MAE-base & 43.55 &  33.60 \\
    \midrule
     \rowcolor{mygray}DiT-B/2 & \ours & 28.26  & 48.92 \\
    \bottomrule
  \end{tabular}
\end{subtable} 
\end{minipage}
\begin{minipage}[t]{0.3\textwidth}
    \begin{subtable}[t]{\linewidth}
\centering
\caption{\textbf{FID of Additional LN}}
\label{tab: ablation_ln}
\setlength\tabcolsep{2pt}
  \begin{tabular}{@{}lcccc@{}}
    \toprule
    Model & 100k & 200k  & 300k & 400k\\
    \midrule
    DiT-B/2 & 40.41 & 33.18 & 30.27 & 28.76 \\
     \rowcolor{mygray}DiT-B/2 + LN & 38.44 & 32.15 & 29.89 & 28.26 \\
    \bottomrule
  \end{tabular}
\end{subtable} 
\end{minipage}
\begin{minipage}[t]{0.3\textwidth}
   \begin{subtable}[t]{\linewidth}
\centering
\caption{\textbf{Pixel Normalization}} %
\label{tab: ablation_pixel_norm}
\setlength\tabcolsep{2pt}
  \begin{tabular}{@{}lcccc@{}}
    \toprule
    Method & Pretrain Loss &SFT Acc &LP Acc &FID\\
    \midrule
    \rowcolor{mygray}\ours &0.465& 82.6\%& 62.8\% & 28.26\\
    - wo PN & 0.289&82.5\%&63.4\%&30.65\\
    \bottomrule
  \end{tabular}
\end{subtable}  
\end{minipage}
\vspace{0.3cm}

\begin{minipage}[t]{0.24\textwidth}
\begin{subtable}[t]{\linewidth}
\centering
\caption{\textbf{
Mask Ratio
} }
\label{tab:abalation_mask_ratio}
  \begin{tabular}{@{}ccc@{}}
    \toprule
    Mask Ratio &SFT Acc & LP Acc \\
    \midrule
    0.40 & 82.7\%& 59.1\%\\
    0.50 & \textbf{82.8\%}& 58.8\% \\
    0.60 & 82.7\% & 62.7\% \\
    \rowcolor{mygray}
    0.75 & 82.6\%& \textbf{62.8}\% \\
    0.85 & 82.2\%& 61.5\%\\
    \bottomrule
  \end{tabular}
\end{subtable} 
\end{minipage}
    \hfill
\begin{minipage}[t]{0.24\textwidth}
    \begin{subtable}[t]{\linewidth}
\centering
\caption{\textbf{Input Resolutions}}
\label{tab:ablation_resolution}
\setlength\tabcolsep{1pt}
  \begin{tabular}{@{}ccccc@{}}
    \toprule
    Resolution &SFT Acc & LP Acc & FID$(\downarrow)$& IS$(\uparrow)$\\
    \midrule
    \rowcolor{mygray}224$\times$224 & 82.6\%& 62.8\% & 28.26& 48.92\\
    256$\times$256 & 83.4\% & 64.3\%& 26.86 & 51.07\\
    \bottomrule
  \end{tabular}
\end{subtable} 
\end{minipage}
\hfill
\begin{minipage}[t]{0.24\textwidth}
    \begin{subtable}[t]{\linewidth}
\centering
\caption{\textbf{Different VAEs.} C: channels, S: down-sampling factor.}

\label{tab:ablation_diff_vae}
\setlength\tabcolsep{1pt}
  \begin{tabular}{@{}lcc@{}}
    \toprule
    VAE &SFT Acc & LP Acc \\
    \midrule
    C4S8 from \cite{peebles2023scalable,rombach2022high} & 82.6\%& 62.8\% \\
    C16S16 from \cite{li2024autoregressive} & 83.2\%& 66.9\% \\
    \bottomrule
  \end{tabular}
\end{subtable} 
\end{minipage}
\hfill
\begin{minipage}[t]{0.24\textwidth}
  \begin{subtable}[t]{\linewidth}
\centering
\caption{\textbf{Layers for Weight Initialization}}
\label{tab: ablation_layer_init}
\setlength\tabcolsep{1pt}
  \begin{tabular}{@{}lc>{\collectcell\hidecontent}c<{\endcollectcell}c>{\collectcell\hidecontent}c<{\endcollectcell}>{\collectcell\hidecontent}c<{\endcollectcell}@{}}
    \toprule
    Method & \textbf{FID} $(\downarrow)$ & \textbf{sFID} $(\downarrow)$ & \textbf{IS} $(\uparrow)$ & \textbf{Pre.} $(\uparrow)$ & \textbf{Rec.} $(\uparrow)$ \\
    \midrule
    DiT-B/2 &42.62 &8.24 &33.67 &0.49 &0.63\\
    \midrule
    \rowcolor{mygray} \ours &28.26 &8.00 &48.92 &0.58 &0.62\\
    - Last 2 Layers &29.63 &8.03 &46.82 &0.57 &0.62\\
    - Last 4 Layers &32.88 &8.11 &42.06 &0.55 &0.61\\
    - Last 6 Layers &37.58 &8.30 &37.89 &0.52 &0.63\\
    \bottomrule
  \end{tabular}
\end{subtable}   
\end{minipage}
\caption{\textbf{Ablation experiments.} Default settings are marked in \colorbox{mygray}{grey}. Additional ablations are provided in Section~\ref{sec:ablation_more}.}
\label{tab:ablations} 
\vspace{-4mm}
\end{table*}

\textbf{Additional Adaptation for LayerNorm.}
The results are presented in Table~\ref{tab: ablation_ln}. Here, “+ LN” denotes that the learnable parameters in LN are utilized, thereby incorporating additional pretrained weights.

\textbf{Pixel Normalization (PN) Target.} We also investigate the impact of per-patch normalization to formulate the target and show the results in  Table~\ref{tab: ablation_pixel_norm}.

\textbf{Mask Ratio.} We sample different mask ratios and report the results in Table~\ref{tab:abalation_mask_ratio}. We choose 0.75 as our default setting because it achieves a good trade-off between training speed and accuracy.

\textbf{Input Resolution.} We further evaluate our approach at an input resolution of 256$\times$256, consistent with the DiT model's default configuration. All models are pretrained for 800 epochs. The results, summarized in Table~\ref{tab:ablation_resolution}, demonstrate that higher resolution leads to improved performance across all tasks. However, to ensure fair comparisons with prior work, we maintain the same training budget and use 224$\times$224 as the default input size.

\textbf{Choice of VAE to Better Image Understanding Performance.} We first use the same VAE as \cite{peebles2023scalable,rombach2022high} to alleviate the distribution mismatch from the pretraining stage to the generation stage. We notice a performance gap between this setting and the original MAE on image understanding tasks. We attribute this gap to limited feature compression and reconstruction of SD-VAE \cite{rombach2022high}. To validate the assumption, we replace it with a VAE of stronger reconstruction in \cite{li2024autoregressive} and report the result in Table~\ref{tab:ablation_diff_vae}.

\textbf{Number of Layers for Weight Initialization.} 
DDAE \cite{xiang2023denoising} demonstrated that the most discriminative features in diffusion models reside in the intermediate layers. Similarly, REPA \cite{yu2025repa} showed that applying representation alignment loss to the first 8 layers improves performance. These findings indicate that early layers primarily capture understanding, while later layers focus on generation. Consequently, as detailed in Table \ref{tab: ablation_layer_init}, we reinitialize the pretrained weights of the last 2, 4, and 6 layers with random weights. Although increasing the number of randomly initialized layers deteriorates generation quality, the performance still exceeds that of the baseline, highlighting the critical importance of pretraining each layer in \ours.

\subsection{Discussion}
We discuss some interesting topics in this section.

\textbf{VAE for Image Understanding.} From the efficient architecture perspective, it's not a perfect idea to apply VAE to designing image backbones for classification. After all, the philosophy of VAE is to pursue higher compression while maintaining stronger reconstruction. The raw image itself is lossless and more efficient. However, it's interesting to see that given a latent representation encoded by a good VAE, we can achieve at least comparable recognition performance.

\begin{figure}[h]
    \centering
    \includegraphics[width=0.9\linewidth]{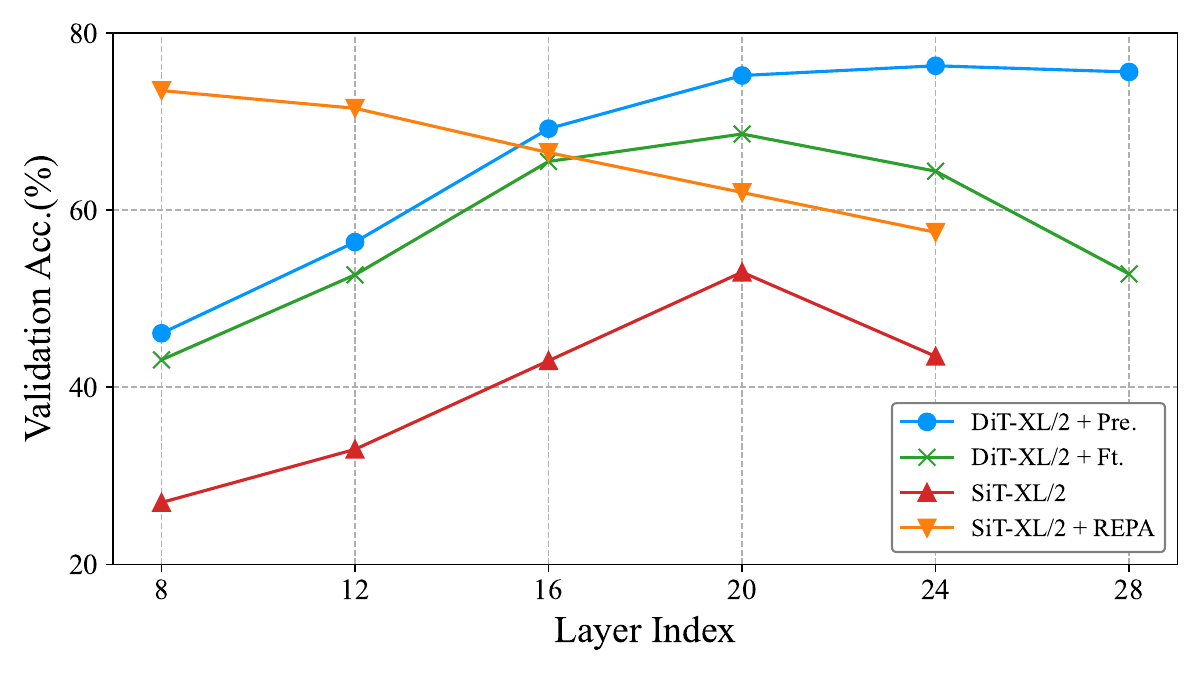}
    \caption{Linear probe on ImageNet across different stages and layers using DiT-XL/2.}
    \label{fig:comp_repa}
    \vspace{-3mm}
\end{figure}

\begin{figure}[h]
    \centering
    \includegraphics[width=1.0\linewidth]
    {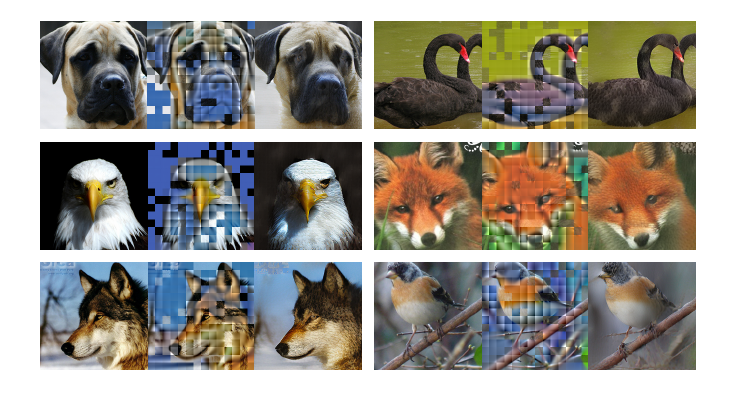}
    \caption{
    \textbf{Image restoration.} (Left) GT, (Middle) MAE restoration with Visible GT patches, (Right) \ours restoration.
}
    \label{fig:image_restore}
\vspace{-2mm}
\end{figure}
\textbf{Working Mechanism.}
As illustrated in Figure \ref{fig:comp_repa}, we compare the linear probing performance of DiT-XL/2 pre- and post-generation training, as well as SiT-XL/2 with and without REPA training. Our findings underscore that effective initialization significantly enhances DiT’s recognition capability, with earlier layers capturing high-level semantic features and later layers optimized for generation, aligning with prior observations \cite{yu2025repa}.
In contrast, REPA employs handcrafted alignment layers, which consistently exhibit superior linear performance (e.g., the 8th layer in Figure \ref{fig:comp_repa}). While these layers demonstrate strong performance, they may not be optimal. Our approach diverges from this methodology by leveraging a carefully curated initialization strategy that enables the network to autonomously identify the most effective layer for linear classification.
Notably, after 400k iterations of generation training using our proposed method, the 20th layer emerges as the optimal layer for linear classification performance. We attribute the enhanced performance of our method, in part, to this innovative mechanism.

\textbf{Alternative Perspective on the Underlying Paradigm.}
We investigate image restoration using ViT-Large models, pretrained for 800 epochs with a mask ratio of 0.75. Both the encoder and decoder are retained to infer masked regions. As depicted in Figure~\ref{fig:image_restore}, our method significantly outperforms MAE in image restoration, underscoring the necessity of robust representations for effective restoration. More results are provided in Figure~\ref{fig:mae_more_results}. This aligns with the diffusion-based generation framework, suggesting that over-tuning the encoder with supervised labels for enhanced discriminative capabilities does not significantly benefit image generation. This may explain the limited success reported in this domain. To empirically validate the hypothesis, we utilized a supervised fine-tuning (SFT) model, which achieved 82.6$\%$ top-1 accuracy on the ImageNet validation set, to initialize the DiT-B/2 model trained for 400k iterations. The results are summarized in Table~\ref{tab:pretrain_sft_init}. Although this configuration outperforms the random initialization baseline, it significantly lags behind the performance of the pretraining group.

\begin{table}[h]
  \centering
  \resizebox{0.8\linewidth}{!}{
  \begin{tabular}{@{}cccc@{}}
    \toprule
    Model Initialization & Top-1($\%$)&FID$(\downarrow)$& IS$(\uparrow)$\\
    \midrule
    random & - & 42.62 & 33.67 \\
    SFT& 82.6 & 39.35 &35.65\\
    \rowcolor{mygray}pretrain & - & 28.26 & 48.92\\

    \bottomrule
  \end{tabular}}
  \caption{SFT tuning severely weakens  generation performances.}
  \label{tab:pretrain_sft_init}
\vspace{-4mm}
\end{table}
\section{Related work}
\label{sec:related_works}
We provide highly related work in the main text and additional related work in Section~\ref{sec:more_related_work}.

\textbf{Self-Supervised Learning.}
Self-supervised learning (SSL) leverages unlabeled data to learn visual representations via pretext tasks. SimCLR \cite{chen2020big} and MoCo \cite{he2020momentum}, contrast positive pairs with negative samples from mini-batches, enhanced by data augmentations. Self-distillation methods like BYOL \cite{grill2020bootstrap} and SimSiam \cite{chen2020simsiam} achieve strong performance without negative samples, aligning representations through self-distillation. Masked Image Modeling (MIM) methods \cite{he2022masked, bao2022beit, beitv2, wei2022masked, vincent2010stacked, chen2020generative} extend denoising autoencoders \cite{vincent2008extracting}, recovering masked regions in images.

\textbf{Diffusion Models.} 
Diffusion models are defined by a process that transforms images into Gaussian noise, followed by a learned reverse denoising process. This can be modeled as stochastic differential equations (SDEs) or ordinary differential equations (ODEs). DDPM \cite{ho2020denoising}  models this process as a Markov chain, while DDIM \cite{song2020denoising} removes the Markov assumption but retains the marginal distribution, accelerating sampling. Flow-based models \cite{flowmatching, liu2022flow, liu2023instaflow} train Continuous Normalizing Flows (CNFs) by learning a vector field for the probability path between distributions and generating samples via ODEs. We validate our method on DDPM-based DiTs and flow-based SiTs.

\textbf{Diffusion Models for Representation Learning.}
Diffusion models have achieved remarkable success in generative tasks \cite{ho2020denoising, Song2020, rombach2022high}, prompting research into their application for representation learning and downstream tasks. Studies \cite{xiang2023denoising, mukhopadhyay2024text, baranchuk2021labelefficient} have shown that discriminative representations from diffusion models outperform traditional self-supervised methods in tasks like image classification and segmentation. Other works \cite{hudson2023soda, chen2024deconstructing, wei2023diffusion} have restructured diffusion models to better suit representation learning, achieving competitive performance in image classification. Additionally, knowledge distillation from diffusion models has been utilized for downstream understanding tasks \cite{yang2023diffusion, li2023your, li2023dreamteacher}. 
While these methods enhance understanding tasks using generative representations, our work aims to develop a unified pretraining approach that improves performance on both generation and understanding tasks.

\textbf{Accelerating Convergence of Diffusion Transformers.}
Our focus is on improving diffusion models without largely altering the architecture, unlike approaches such as \cite{chu2024visionllama, vavae}, which introduce architectural changes to enhance performance. Given this limitation, the core idea of accelerating the convergence of diffusion models is to leverage visual representations
\cite{gao2023masked, Zheng2024MaskDiT, yu2025repa, vavae, reals}. Most approaches achieve this by adding various types of losses during the training of diffusion models \cite{Zheng2024MaskDiT, yu2025repa}. For instance, REPA \cite{yu2025repa} aligns the representations of diffusion models with external visual representations, enhancing both convergence speed and generation quality. Similarly, studies such as VA-VAE \cite{vavae} and ReaLs et al. \cite{reals} propose aligning representations during VAE training to address the lack of clear semantic representations in traditional VAE latents. 
MaskDiT \cite{Zheng2024MaskDiT} adds an MAE decoder to DiT and incorporates a masked token reconstruction objective during training, improving the training efficiency of diffusion models. MDT \cite{gao2023masked} performs reconstruction on all image tokens and introduces carefully designed modules to the original DiT architecture, achieving better generation performance despite the increased training cost.
While these works focus on accelerating the convergence and improving the performance of diffusion models, our method decouples pretraining from downstream tasks such as image generation and recognition. This allows our approach to offer a more unified and efficient solution.

\section{Conclusion}
In this work, we delve into the pretraining-fine-tuning paradigm, targeting both image recognition and generation tasks. By conducting masked latent modeling within the latent space of a VAE, we achieve the learning of unified, robust representations. This approach necessitates only a single training phase, after which the model can be seamlessly integrated into downstream tasks via weight initialization, thereby yielding substantial performance improvements. Notably, our method incurs no additional training costs or GPU memory overhead for downstream tasks. We posit that our approach serves as a potent baseline for the research community.

\clearpage

{
    \small
    \bibliographystyle{ieeenat_fullname}
    \bibliography{main}
}

\clearpage
\clearpage
\setcounter{page}{1}
\maketitlesupplementary

\appendix

\section{More Ablation Studies}\label{sec:ablation_more}

\textbf{Incorporating Noise into the Pretraining Stage.} We investigate the effect of introducing noise into the latent space during pretraining by adopting the formulation \(x_t = \sqrt{\bar{\alpha}_t} x_0 + \sqrt{1 - \bar{\alpha}_t} \epsilon\), with the objective of reconstructing the unmasked clean target. Despite the similarity of this setting to the data distribution used in generative tasks, it yields a FID of 32.20 and an Inception Score (IS) of 43.36 without Classifier-Free Guidance (CFG), which is inferior to our baseline performance. This suggests that masking modeling already serves as a highly effective form of data augmentation, and the introduction of additional, stronger noise significantly increases the difficulty of the learning task.




\textbf{High-Resolution Results} 
Table~\ref{tab:comp_512} demonstrates the effectiveness of \ours at a higher resolution of 512$\times$512. We initialize the downstream image generation task using weights obtained from pretraining with \ours at 256 resolution and directly transfer them to the 512 resolution setting, adjusting the positional encodings via bilinear interpolation. The results confirm that \ours maintains strong performance and transferability under higher-resolution settings.

\begin{table}[h]
    \centering
    \resizebox{0.9\linewidth}{!}{
    \begin{tabular}{l c c c c }
        \toprule
        \textbf{Model} & \textbf{Params} & \textbf{Steps} & \textbf{FID} $(\downarrow)$ & \textbf{IS} $(\uparrow)$ \\
        \midrule
        SiT-B/2 & 130M & 400K & 42.80 & 37.37 \\
        \rowcolor{mygray}\ours & 130M & 400K & 33.89 & 45.03 \\
        \bottomrule 
    \end{tabular}
    }
        \caption{Results at 512$\times$512 Resolution.}  
    \label{tab:comp_512}
\end{table}

\textbf{Image Normalization (IN).} Image normalization is a standard transformation in the community, and the default mean ([0.485, 0.456, 0.406]) and std ([0.229, 0.224, 0.225]) are widely used. However, VAE of DiT \cite{peebles2023scalable} has a different setting (mean=[0.5, 0.5, 0.5], std=[0.5, 0.5, 0.5]). We perform pretraining using these two groups and report the downstream results in Table~\ref{tab:ablation_image_norm}. Although the default setting of ImageNet has a lower loss (0.375), it doesn't bring in higher performance. Therefore, we utilize the settings of SD-VAE \cite{rombach2022high}.

\begin{table}[h]
  \centering
  \small
  \begin{tabular}{@{}lcccc@{}}
    \toprule
    Method & $Loss_{pretrain}$ &$Acc_{SFT}$ & $Acc_{LP}$ &FID\\
    \midrule
    \rowcolor{mygray}\ours &0.465& 82.6\%& 62.8\% & 28.26\\
    ImageNet IN & 0.375& 82.0\% & 60.8\%&78.23\\
    \bottomrule
  \end{tabular}
  \caption{All results are reported using the same VAE as \cite{peebles2023scalable}.}
  \label{tab:ablation_image_norm}
\end{table}

\textbf{AdaLN-Zero or Skip Connection.} AdaLN-zero initializes the attention and MLP branches with zero weights, effectively disabling them at the beginning of training. This approach alleviates the difficulties associated with the training of deep transformers \cite{peebles2023scalable}. We explored an alternative initialization strategy where the attention and MLP blocks are activated from the start by calibrating the gate bias to 1. This setting achieved similar performance to the zero-initialized approach.  However, considering the minimal modification required for DiT and the established effectiveness of AdaLN-Zero in stabilizing training, we opted to retain the original AdaLN-Zero initialization scheme.

\textbf{Comparison with UMD.} UMD \cite{hansen2024unified} integrates the diffusion loss and MAE loss through a weighted sum approach, aiming to achieve robust performance in both understanding and generation tasks. However, it still significantly underperforms compared to its single-task counterparts in each task. We attribute this shortfall to inherent conflicts between the MAE and diffusion models arising from their coupling.
 
 In terms of performance and efficiency, our method achieves a substantial reduction in training cost, requiring only $15\%$ of the computational resources compared to the DiT baseline to match its performance (see Table~\ref{tab:training_cost}). This highlights the superior efficiency of our approach. In contrast, UMD \cite{hansen2024unified} significantly underperforms the DiT baseline under the same computational constraints, further corroborating the effectiveness of our method. For image recognition tasks, our method achieves performance that is on par with, and in some cases surpasses, the strong baseline MAE. By contrast, UMD falls significantly short, demonstrating a clear performance gap. The results reported in Table 3 of the UMD study are not reliable and significantly deviate from the commonly reproduced and published results in the community. For instance, the FID score for DiT-L/2 with 400 epochs is reported as 9.6, which is consistent with widely accepted results. In contrast, it is known that the DiT-XL/2 architecture requires approximately 1400 epochs to achieve a similar FID score. 

\section{More Related Work} \label{sec:more_related_work}

\textbf{Generative Models with Auxiliary Task.}
MaskDiT \cite{Zheng2024MaskDiT} introduces an asymmetric encoder–decoder architecture based on the DiT framework, leveraging masked reconstruction to reduce the training cost of diffusion models. However, this approach involves substantial modifications to the original DiTs, resulting in limited transferability, and because the encoder always receives noisy inputs, it cannot be applied to downstream recognition tasks.
Similarly, MDT \cite{gao2023masked} employs an additional decoder for mask token modeling to enhance semantic contextual learning. Unlike \cite{Zheng2024MaskDiT}, it performs noise prediction on all tokens rather than just on the unmasked ones. Although this improves generation quality, it also introduces significant computational overhead. Essentially, these methods incorporate an extra masked reconstruction task alongside noise prediction, which compromises architectural flexibility and, due to the input mismatch, restricts their applicability to understanding tasks.
MAGE \cite{li2023mage} proposes a unified framework for image generation and self-supervised representation learning, simultaneously conducting generation and representation learning through a variable mask ratio and an additional contrastive loss. MAGE utilizes VQ-GAN \cite{esser2021taming} encoder and quantizer to tokenize the input images and focuses solely on class-unconditional generation, whereas our approach operates in continuous space, aiming to enhance the generation performance of diffusion models while maintaining strong representation.
In contrast, our method introduces minimal modifications to the original DiT/SiT architecture, ensuring excellent transferability and scalability. Moreover, by employing a single masked token reconstruction task, we decouple the heterogeneous optimization objectives between pretraining and downstream tasks.

\textbf{MLLMs for Unified Understanding and Generation.}
Multimodal Large Language Models (MLLMs) have recently drawn extensive attention from both academia and industry. MLLMs \cite{qwen-vl, internvl25, lu2024deepseekvl} enable visual question answering in multimodal understanding tasks by aligning image embeddings with textual embeddings and jointly feeding them into a large language model, ultimately yielding text token IDs.

Several works address unified multimodal understanding and generation tasks, which can be broadly divided into two categories. One line of research \cite{team2024chameleon, wu2024janus, xie2024show-o, lu2022unified-io} employs VQ-GAN \cite{esser2021taming} or VQ-VAE \cite{vqvae} to tokenize images into discrete token IDs that are then fed into MLLMs for autoregressive image generation, thus aligning with the discrete input format of large language models. To mitigate potential performance degradation on understanding tasks due to discretization, \cite{wu2024janus} proposes utilizing discretized image token IDs only during the image generation stage.
Another line of research \cite{ma2024janusflow, zhou2024transfusion, gupta2022metamorph, ge2024seedx, dong2024dreamllm} does not require converting images into discrete token IDs consistent with text. Instead, it leverages the image tokens output by the LLM as conditions for external generative models (e.g., Stable Diffusion \cite{LDM}) to produce images. Our approach is a purely vision-based pre-training method, providing a robust weight initialization for subsequent fine-tuning on downstream understanding and generation tasks.

\section{Visualization}

\begin{figure*}[h]
  \centering
  \includegraphics[width=\linewidth]{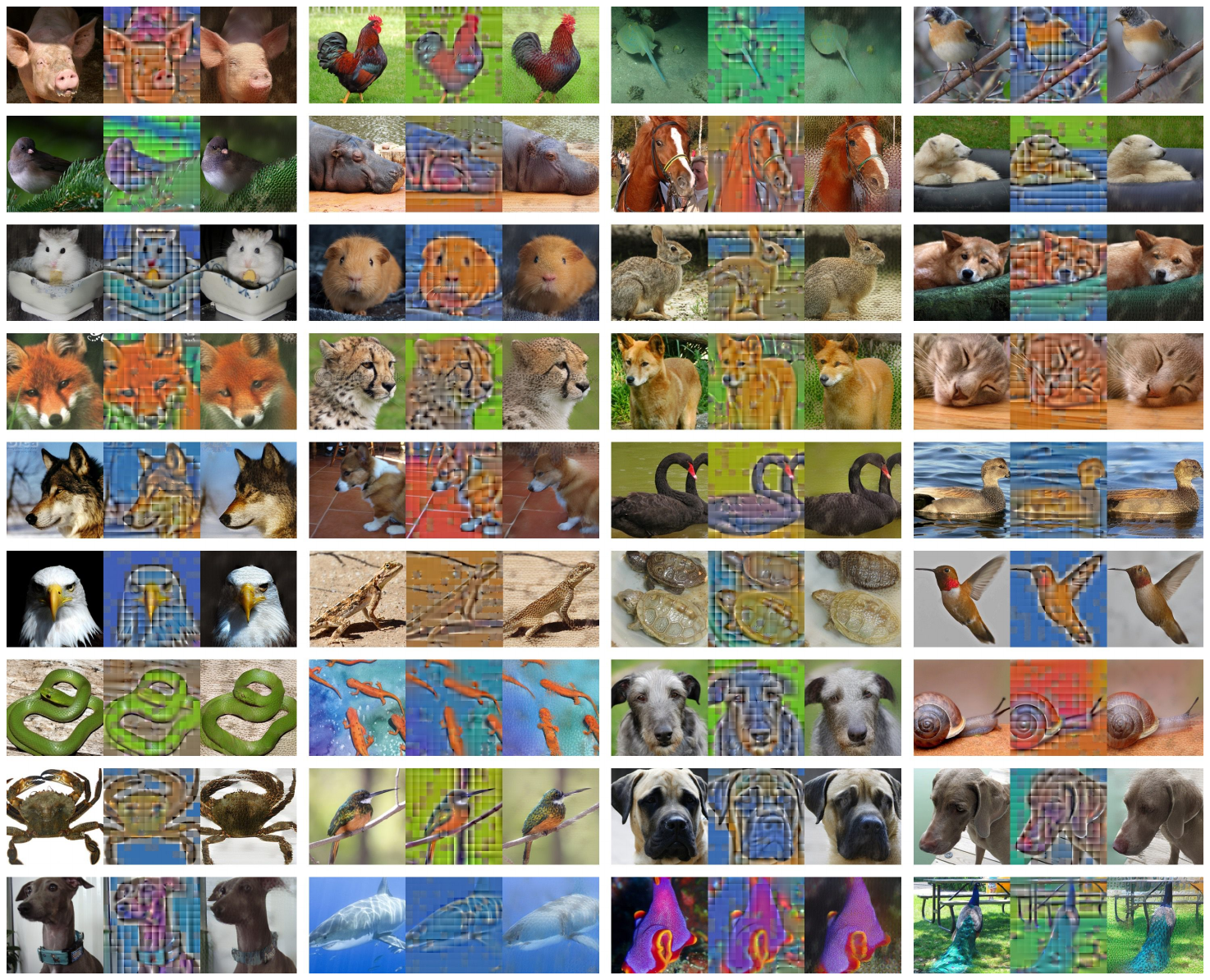}
  \caption{
  Reconstruction results using ViT-Large on the ImageNet validation set. For each group of samples, we present the ground-truth image (left), MAE \cite{he2022masked} (middle) reconstructed image and \ours reconstructed image (right). The masking ratio is set to 75$\%$. 
}
  \label{fig:mae_more_results}
\end{figure*}

\begin{figure*}[h]
  \centering
  \includegraphics[width=0.8\linewidth]{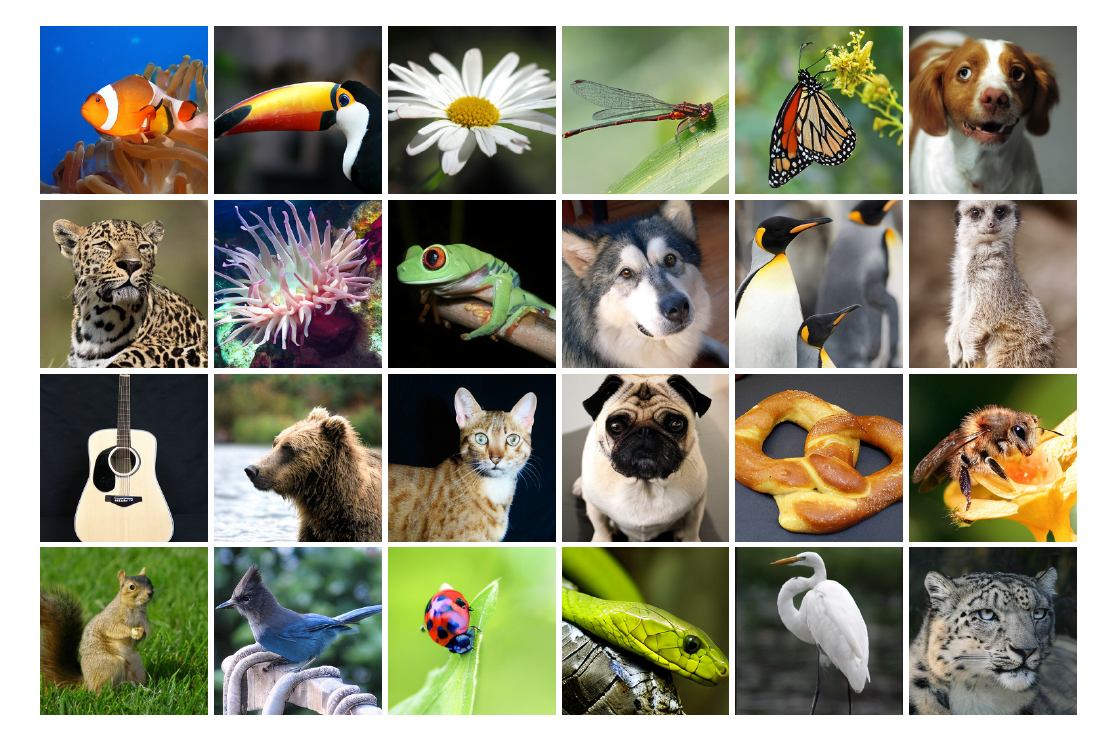}
  \caption{
  256$\times$256 generation samples: DiT-XL/2 (1.2M steps) with CFG=4.0.
}
  \label{fig:dit_more_results}
\end{figure*}

\subsection{Image Restoration}

We visualize image reconstruction results using a ViT-Large model (encoder + decoder) pretrained with MAE and our method (see Figure~\ref{fig:mae_more_results}). We randomly mask $25\%$ of the patches and infer the restored images. Our method achieves much better performance.

\begin{figure*}[h]
  \centering
  \includegraphics[width=0.8\linewidth]{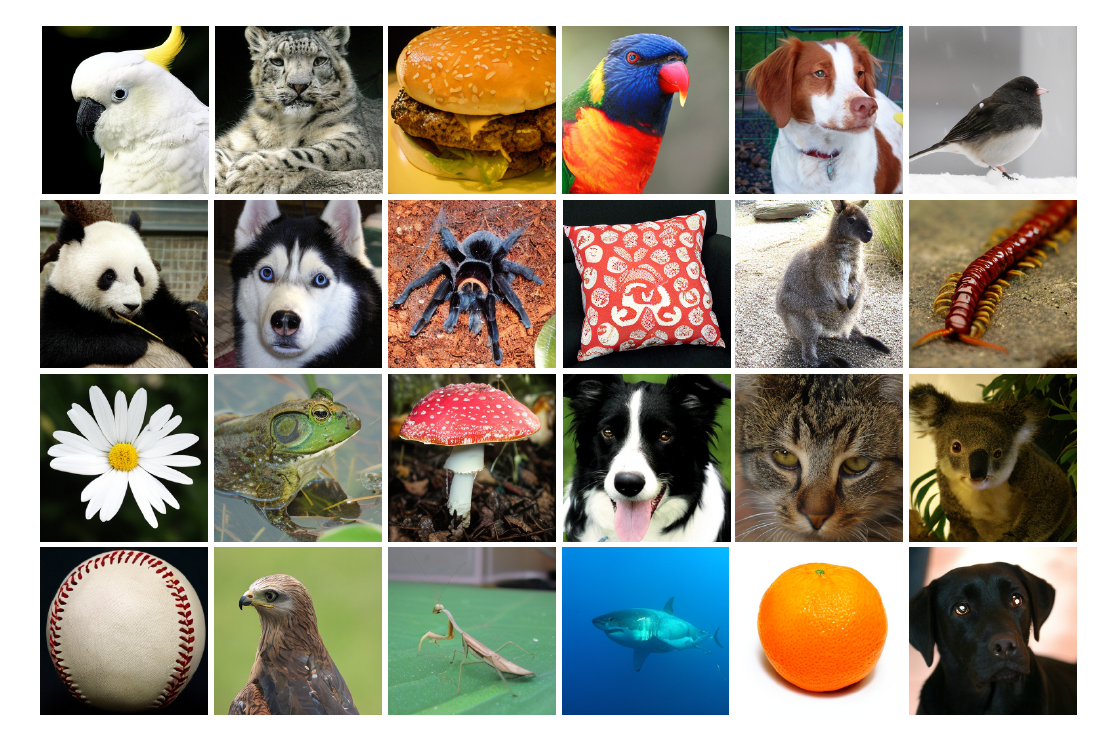}
  \caption{
  256$\times$256 generation samples: SiT-XL/2 (800K steps) with CFG=4.0.
}
  \label{fig:sit_more_results}
\end{figure*}

\begin{table}[ht]
\centering
\begin{tabular}{ r c}
\toprule
config & value \\
\midrule
optimizer & LARS \cite{you2017large} \\
base learning rate & 0.1 \\
weight decay & 0 \\
optimizer momentum & 0.9 \\
batch size & 16384 \\
learning rate schedule & cosine decay \\
warm-up epochs & 10 \\
training epochs & 90(B), 50(L) \\
augmentation & RandomResizedCrop \\
\bottomrule
\end{tabular}
\caption{\textbf{Linear probing setting.}
\label{tab:hyper_mae_linear}}
\end{table}

\subsection{Fully Tuning on ImageNet}
The hyperparameter setting for fine-tuning in ImageNet is shown in Table~\ref{tab:mae_finetune}.

\subsection{Image Generation on ImageNet}
The hyperparameter setting for generation in ImageNet is shown in Table~\ref{tab:img_gen}.

\begin{table}[ht]
\centering

\begin{tabular}{ r c}
\toprule
config & value \\
\midrule
optimizer & AdamW \\
base learning rate & 1e-3 \\
weight decay & 0.05 \\
optimizer momentum & $\beta_1, \beta_2{=}0.9, 0.999$ \\
layer-wise lr decay \cite{bao2021beit} & 0.75 \\
batch size & 1024 \\
learning rate schedule & cosine decay \\
warmup epochs & 5 \\
training epochs & 100 (B), 50 (L) \\
augmentation & RandAug (9, 0.5) \cite{cubuk2020randaugment} \\
label smoothing \cite{szegedy2016rethinking} & 0.1 \\
mixup \cite{zhang2017mixup} & 0.8 \\
cutmix \cite{yun2019cutmix} & 1.0 \\
drop path & 0.1  \\
\bottomrule
\end{tabular}
\caption{\textbf{Fine-tuning the whole neural network.}}
\label{tab:mae_finetune} 
\end{table}

\begin{table}[h]
\centering

\begin{tabular}{c r c}
\toprule
model & config & value \\
\cmidrule(lr){1-3}

\multirow{6}{*}{DiTs}
&optimizer & AdamW \\
&constant learning rate & 1e-4 \\
&weight decay & 0. \\
&optimizer momentum & $\beta_1, \beta_2{=}0.9, 0.999$ \\
&batch size & 256 \\
&augmentation & RandomHorizontalFlip \\

\cmidrule(lr){2-3}

\multirow{9}{*}{SiTs}
&optimizer & AdamW \\
&constant learning rate & 1e-4 \\
&weight decay & 0. \\
&optimizer momentum & $\beta_1, \beta_2{=}0.9, 0.999$ \\
&batch size & 256 \\
&augmentation & RandomHorizontalFlip \\
&path type & Linear \\
&prediction & velocity \\

\bottomrule
\end{tabular}
\caption{\textbf{Image generation on ImageNet.}}
\label{tab:img_gen} 
\end{table}

\subsection{Image Generation}\label{sec:image_gen_256}
We visualize more image generation results from DiT-XL/2 and SiT-XL/2, as shown in Figure \ref{fig:dit_more_results} and Figure \ref{fig:sit_more_results}, respectively. All results are generated with a CFG scale of 4.0.

\section{HyperParameters}

\subsection{Linear Probe on ImageNet}
We follow the setting of \cite{he2022masked, chu2024visionllama} and show the details in Table~\ref{tab:hyper_mae_linear}.

\section{Pretraining Code}
We provide the code for the pre-training stage bundled with the supplementary materials. The pre-trained weights can be conveniently transferred to downstream understanding and generation tasks.

\end{document}